# NaturalTurn: A Method to Segment Speech into Psychologically Meaningful Conversational Turns

Gus Cooney & Andrew Reece

**Abstract**

Conversation is a subject of increasing interest in the social, cognitive, and computational sciences. Yet as conversational datasets continue to increase in size and complexity, researchers lack scalable methods to segment speech-to-text transcripts into conversational “turns”—the basic building blocks of social interaction. We discuss this challenge and then introduce “NaturalTurn,” a turn-segmentation algorithm designed to accurately capture the dynamics of conversational exchange. NaturalTurn operates by distinguishing speakers’ primary conversational turns from listeners’ secondary utterances, such as backchannels, brief interjections, and other forms of parallel speech that characterize human conversation.
Using data from a large conversation corpus, we show that NaturalTurn captures conversational turns more accurately than a baseline model. For example, it produces turns with durations and gaps that match empirical literature, reveals stronger linguistic alignment patterns between speakers, and uncovers otherwise hidden relationships between turn-taking and affective outcomes. NaturalTurn thus represents a pragmatic development in machine-generated transcript-processing methods, or “turn models”, that will enable researchers to link turn-taking dynamics with important outcomes of social interaction, a central goal of conversation science.

Humans are extraordinarily sensitive to the timing and exchange of conversational turns. This nuanced sensitivity emerges early in childhood [1, 2], is remarkably stable across cultures [3], and can be impaired in a range of clinical and developmental disorders [4-6]. Strict adherence to the unspoken rules of turn-taking is what makes conversation such a powerful device for coordinating minds [7-10]. As such, a thorough understanding of conversation, together with any social outcomes that result, hinges on accurately identifying whose turn it is at each moment in a conversation, ideally a close approximation of the "naturalistic turns" experienced by the actual participants in the dialog.

The ease with which most people navigate turn-taking might suggest that segmenting a conversation into turns would be relatively straightforward. Whether at a company meeting or a dinner party, participants discern the current speaker intuitively and can easily separate true turns from brief overlapping speech or ancillary interjections. However, codifying these intuitions into a formal sequence of instructions for a machine to follow is a formidable task. The main obstacle is that human conversations rarely unfold like orderly walkie-talkie exchanges in which only one person talks at a time and every turn ends with a definitive "over" or "out" cue [9]. Instead, conversation is frequently a riot of overlapping speech, in which listeners constantly interweave brief interjections that clarify, affirm, and coordinate the dialogue [11-13]. To a machine, these unpredictable overlaps represent a tangle of confounding edge cases, particularly when multiple individuals speak simultaneously within a shared conversational space. To date, few open-access turn-segmentation algorithms exist. Rarer still are algorithms that can produce useable results without heavy human annotation post-transcription.

This scarcity of methods has significant implications. As early as the 1940s, scholars recognized that turn-taking dynamics might encode rich social information, revealing personality traits and interpersonal patterns that could predict outcomes like personnel hiring [14, 15]. This early promise sparked decades of tantalizing findings: researchers discovered synchrony in conversational rhythms [16, 17], linked turn duration to impression formation [18], and showed how turn-taking speed predicts whether conversation partners "click" and enjoy one another's company [19]. Yet despite these glimpses of turn-taking's social significance, the field has never fully realized its potential, leaving us with isolated findings rather than systematic understanding. This persistent gap stems, at least in part, from the lack of automated methods to segment transcripts into turns at scale.

Accurate turn segmentation algorithms are also particularly important in today's era of generative-AI-assisted analytics using large language models (LLMs). Critically, researchers may provide LLMs large batches of text and then prompt the models to code conversations at the turn level (e.g., "What is the dominant emotion for every turn in this corpus?", or "What is the semantic distance between this turn and the previous?"). However, without a well-justified definition of what constitutes a turn, LLMs will operate on whatever turn structure they are given, a classic "garbage in, garbage out" problem in machine learning. Moreover, the varied and often unexamined definitions of turns produced by different disciplines and transcription pipelines introduce considerable "researcher degrees of freedom," potentially undermining the replicability and robustness of the science. As sense-making in human interaction data is increasingly outsourced to AI, the attention given to upstream data collection and preprocessing methods—such as segmenting transcripts into conversational turns—becomes more important than ever.

Thus, a central issue for researchers is that most widely-used speech-to-text services (e.g., Amazon Web Services (AWS) Transcribe, Microsoft Azure Speech-to-Text, OpenAI's Whisper, etc.) excel at recovering words from audio, yet they deliver transcripts with inaccurate turns because they conflate the speaker's primary contribution with listener's feedback. The result is a stream of micro-segments whose boundaries rarely match the units that conversational partners themselves experience as "turns" (Figure 1). Researchers studying conversation and social interaction at scale therefore face a bottleneck: either accept these fragmented transcripts or invest in bespoke post-processing. A variety of sophisticated turn-taking models has been proposed—ranging from prosody-based classifiers to neural network forecasters—but many of them target real-time dialogue management rather than corpus-level segmentation, and they typically require annotated training data, specialized features, or heavy computation. In the following section, we review these developments in detail and show how our contribution, NaturalTurn, differs.

NaturalTurn is a deterministic, annotation-free algorithm that partitions an automatic speech recognition (ASR) transcript into naturalistic turns; that is, stretches of talk that the participants themselves would have experienced as one person "holding the floor." NaturalTurn attempts to identify each speaker's primary floor-holding segment and separate out all secondary vocalizations: canonical backchannels (e.g., "mhm," "yeah"), brief affective interjections (e.g., "oh wow"), and other parallel speech that listeners produce while someone else is speaking. These secondary events are preserved with their original timing but are stored alongside, rather than inside, the primary-turn timeline. Using heuristic-based architecture, NaturalTurn represents an initial, largely pragmatic solution to a complex challenge, one that will benefit from additional development from multidisciplinary research teams. Nevertheless, with only a few easily-tunable parameters and no additional feature annotation or machine learning, NaturalTurn can be added to any transcription pipeline and scaled to millions of turns, giving researchers a psychologically plausible unit of analysis for a wide range of conversational studies.

In the sections that follow, we explain NaturalTurn's algorithm design and functionality. We provide evidence that our algorithm outperforms the "baseline" transcripts generated by modern speech-to-text services by improving the representation of two foundational attributes of turn-taking: turn durations and the timing of turn exchanges. After detailing these mechanics, we provide a case study using NaturalTurn on a large public corpus of dyadic conversations accompanied by people's post-interaction ratings of rapport, enjoyment, and impression formation. These results demonstrate the practical benefits of our method and pave the way for novel research connecting the dynamics of turn-taking with broader social outcomes. Finally, we discuss the value of our algorithm as a contribution to a broader class of machine-generated transcript-processing methods, or "turn models," that will be indispensable for the evolving interdisciplinary science of conversation.

## Background and Related Work

Researchers in many disciplines have tried to segment conversational speech into units larger than a word. The most common unit is the "turn," yet what counts as a turn varies considerably across time, subfield, and application. In this section, we briefly review some major perspectives from the domains of conversation analysis, corpus and computational linguistics, phonetics, dialogue-system engineering, and psychology, to locate our contribution, NaturalTurn. The NaturalTurn algorithm is deliberately pragmatic: it segments speech into psychologically plausible turns *at scale* using only the text and timing delivered by a modern speech-to-text service. Built as an easily inserted pre-processing layer, it is meant to help researchers work with large conversational data sets without first mounting a bespoke annotation effort—especially researchers interested in linking turn-by-turn dynamics to social outcomes.

## Approaches to Turn Segmentation

***Conversation Analysis.*** The foundational account of turn-taking comes from Conversation Analysis, hereafter called CA. Sacks et al. [20] argued that talk is organized by a "simplest systematics" that lets speakers locally coordinate *who* talks, *when*, and *for how long*, with remarkably few gaps or overlaps. CA introduced two concepts that still frame the discussion: *turn-constructional units* (TCUs), which are building blocks such as clauses, phrases, or even single words; and *transition-relevance places* (TRPs), which are points at which speaker change becomes possible. A turn can comprise one or many TCUs, and multiple TRPs may appear before the "floor" is actually turned over. What sets CA apart as a field is the concept that turns are not just stretches of talk; rather, they are social actions, often part of larger sequences of turns that perform specific functions such as questioning, requesting, or offering [21]. In this view, even a single word (or gesture) can constitute a turn if it completes a social action. Alternatively, a single speaker may produce a long multi-unit turn such as when narrating a story or explaining something at length. In practice, CA relies on careful human transcription rather than fixed formal rules, creating challenges for performing scalable, automated turn detection and analysis.

***Early computational approaches to segmentation.*** Early computational work on large corpora used the one cue that early-generation computers could reliably detect: who is audible and who is silent [22-24]. In a dyadic conversation, a voice-activity detector (VAD) chops each speaker's audio channel into *inter-pausal units* (IPUs), which are stretches of speech bounded by silences longer than a hand-tuned threshold (e.g., 250 ms). This silence threshold determines what counts as a gap between speech segments. Once IPUs are in hand, researchers might make use of two simple rules for finding turns. The *silence-threshold rule* ends a turn whenever the current speaker stays quiet longer than the gap value, even if the partner says nothing. The *channel-switch rule* closes the current turn the instant an IPU appears on the other channel, grouping a speaker's consecutive IPUs into the same turn only so long as no cross-channel IPU intervenes. Both rules remain common in transcription pipelines, yet they often fail to capture typical features of human conversation. First, a silence need not mark completion of a turn, as speakers often pause mid-utterance to think, breathe or respond to something in their environment; when these pauses exceed the silence threshold, they are misidentified as turn boundaries. Second, listeners frequently produce short verbal overlaps without intending to "take the floor." In the transcript text, "*I was standing to the [listener: mhm] left of the … [speaker: pause, gaze averted, hand raised] … brown building,*" a pure VAD-based rule might insert spurious boundaries at the listener's "mhm" or at the speaker's multimodal hesitation, fragmenting what the participants themselves experienced as a single turn.

***Beyond uninterrupted speech and silence***. Many fields have tried to move beyond defining turns as uninterrupted stretches of speech between silence thresholds by layering in additional cues to decide where a turn really ends. In language-acquisition work, for example, the *T-unit* and *C-unit* conventions segment speech at syntactic and semantic boundaries, to ensure that each turn is a clause-complete thought [25]. Fields related to phonetics and acoustics added intonation criteria, considering that a long pause might only be a boundary if it was preceded by a falling tone, whereas a "half-complete" contour signaled continuation [26, 27]. Early work in psychology by Duncan et al. [28-31], as reviewed by Rühlemann and Gries [32], attempted to catalogue an even richer set of behaviors including vowel lengthening, loudness decay, gesture cessation, and gaze return. Applied to our running example—"*I was standing to the [listener: mhm] left of the … [pause, gaze averted, hand raised] … brown building*"—these additional turn-related cues could be used to write additional rules that ignore the listener's *"*mhm" (short, rising classic backchannel), treat the ensuing pause as mid-turn (no final lowering, gaze still averted, hand signaling "hold on"), and wait for the pitch fall and final-syllable lengthening after *"building*" before declaring a turn boundary. Although potentially predictive, each extra cue comes with trade-offs in sensitivity, specificity and complexity of design.

***Multi-cue snapshot models.*** Having combined cues from syntax, semantics, prosody, gaze, and gesture, researchers next began to explore *integrating* many signals at every prospective boundary. In one study, Gravano and Hirschberg [33] extracted many lexical, syntactic, and acoustic–prosodic features from the final 200–500 ms before each verbal pause and trained shallow classifiers to decide whether the upcoming gap would be a "hold" or a "switch". The identical framework, with a different cue subset, was even able to predict backchannel-inviting positions. A key finding was an *additive* effect: each extra turn-yield cue—say, clause completion plus falling F0 plus final-syllable lengthening—raised the switch probability, whereas clusters of turn-holding cues such as continuing intonation or gaze aversion favored continuation [34].

Such research relies on hand-curated snapshots: a fixed bundle of features selected by the researcher and measured in a brief window before each pause. This is often adequate for offline annotation; however, turn-segmentation research is today driven by those interested in building real-time dialogue systems, as discussed below. For these applications, treating every boundary in isolation and using features only from a short slice of time can miss evolving trends in a conversation, and more seriously, can leave a machine unable to start acting until *after* a silence has begun, which is too late for seamless turn exchange. This intolerable latency drove the field toward continuous-sequence models that monitor the full audio signal in real time and attempt to predict a turn change before the silence emerges, giving the system the head-start it needs.

***Continuous-sequence models.*** After snapshot classifiers, dialogue-system research moved to rolling forecasters. A typical rolling forecaster is a neural network that updates every few dozen milliseconds and predicts whether the speaker will still hold the floor a few hundred milliseconds later [35]. The data used to train these models are often labelled automatically: whenever a silence exceeds some threshold (e.g., 250 ms) the next stretch of speech is tagged HOLD if the same speaker resumes, or SWITCH if the partner begins. As an example, a Long Short-Term Memory (LSTM) recurrent neural network ingests frame-level acoustics and the stream of words (along with other ASR outputs like embeddings, Part-of-Speech labels, etc.), and learns this binary sequence-labeling task. At run-time, it emits a “probability of switch” curve. A similar processing loop still underlies rolling forecaster systems today; however, later studies [36-39] broadened the training data with richer ASR word streams, added extra cues for dialogue-act and filler detection, and split the model into fast acoustic and slower lexical pathways. These advancements improved model performance in detecting not only brief backchannels but also large overlaps, long pauses, and other complex hand-off cases.

Today, turn-taking systems have moved beyond simple LSTM pipelines to transformer-based architectures (e.g., TurnGPT, the Voice-Activity Projection (VAP) family), often using more advanced speech encoders (e.g., Conformer, wav2vec 2.0, HuBERT). Many of these models fuse multiple cues (acoustics, lexical context, sometimes gaze) and often are able to generate frame-by-frame voice-activity forecasts that indicate who is likely to speak at each upcoming moment. Because the model sees a longer context and predicts a detailed timeline, researchers can extract fine-grained phenomena (smooth switches, overlaps, backchannels) from the forecast, while system builders gain a unified signal for turn-timing, barge-in handling, and overlap management with minimal task-specific tuning.

Because this literature now evolves monthly and remains focused on live-agent timing, we refer readers to recent surveys [40, 41] and stay focused on our offline goal: transparent, corpus-ready turn segmentation that approximates how participants themselves parcel the conversation—no annotation or neural training required.

***Multiclass offline labelers.*** It is worth noting that in parallel to research efforts to develop real-time turn-switching prediction, a second line of work—much closer to our aims—runs similar models *after* a conversation has finished, when latency no longer matters. With access to the full dialogue, these models can look both forward and backward and replace the coarse *hold/switch* tag with a richer taxonomy of events. Brusco and Gravano [42], for example, used a bidirectional LSTM with long-range cues, such as how long each speaker has held the floors to classify every gap or overlap as a smooth switch, latching continuation, competitive or failed interruption, or backchannel, among other categories. Ekstedt and Skantze [43] slid a broad context window across the audio and turned the resulting probability tracks into labels such as shift-at-silence, shift-during-speech, imminent backchannel, or hold. Boudin et al. [44] extended this offline work from "who-takes-the-floor next?" to focus more directly on "what does the listener actually do?" Running a model over acoustics, words, smiles, nods, and laughter, they located every feedback token and then labeled each as a either a generic "mhm"-style acknowledgement or specific sub-types of listener feedback. Freed from real-time constraints, these offline systems yield conversation-analytic annotations that are immediately useful to researchers, at the cost of requiring the entire recording before any labels can be produced.

***Summary.*** These previous research arcs share many cues, methods, and terminology, but pursue different goals. NaturalTurn therefore does not compete with the latest transformer forecaster or attempt to supply highly granular dialogue-act labels. Instead, it offers a simple, annotation-free, corpus-ready segmentation (primary turns plus secondary listener feedback) that researchers can apply to any transcript to obtain a consistent, socially meaningful unit of analysis, especially useful for linking turn dynamics to psychological and relational outcomes.

## Our Motivation, Approach, & Contribution

Despite more than eighty years of scholarship, empirical work on the social consequences of the turn-taking system remains uneven. A complete survey is provided elsewhere [45]; here we sketch several examples of the central promise and the persistent bottlenecks that formed the motivation for NaturalTurn.

First, a long tradition links turn-duration patterns to rapport, status, personality, and interactional success. Studies range from Chapple's "interaction chronograph" work of the 1940s [14, 15, 46], through clinician-patient research showing duration-matching behavior, that "little questions get little answers" [47-49], to duration synchrony analyses of conversational speech [16, 50], and the macro-rhythm effects on interaction quality [51]. Impression-formation findings reveal that speakers who monopolize time emerge as leaders [18, 52] but are not necessarily the most liked [53, 54]. Although provocative, these findings remain intermittent and under-explored. Technological hurdles related to turn segmentation are at least partly to blame. Studies relying on idiosyncratic, often non-naturalistic turn definitions leave the field without a shared, replicable unit of analysis. Robust results cannot accumulate and theory-building stalls.

A related pattern emerges in research on the timing of turn exchange. Unlike the intermittent literature on duration, the mechanics of "handing over the floor" have been studied in detail, but primarily from a cognitive standpoint, such as the millisecond-level cognitive processes that allow interlocutors to predict and coordinate turn transitions [9, 55, 56]. Only recently, however, have social scientists asked how the nature of these intervals (the magnitude of gaps and overlaps between turns) relates to subjective rapport, "clicking," and conversational enjoyment [19, 57]. As this line of work expands, progress will require a standardized, psychologically natural method for segmenting turns, so that the gaps and overlaps being measured truly correspond to boundaries that participants themselves experience.

Listener backchannels illustrate a third example of research that is limited by turn segmentation. The floor-management function of backchannels is well documented: generic continuers such as "mhm" and "yeah" help speakers maintain the floor and coordinate exchanges [13, 58-60]; yet large-scale and experimental evidence for their social consequences remains underexplored. The clearest demonstrations come from targeted experiments on storytelling. While generic continuers merely signal "keep going," specific tokens such as "Exactly!" or "Wow!" actively shape narrative structure and enhance perceived interaction quality [11, 61]. Unfortunately, most automated speech-to-text pipelines treat every channel change as a new "turn," so backchannels are misclassified as full floor grabs, obscuring the very feedback behavior that researchers might analyze at scale.

In contrast, research on phenomena related to linguistic and behavioral alignment has flourished. Dozens of studies show that interlocutors converge in pronunciation, vocabulary, pitch, gesture, gaze, and emotion, and that this coordination predicts rapport and interactive success [62-70]. Yet many of these findings rest on whatever "turns" a speech-to-text service happens to produce or on idiosyncratic, researcher-defined boundaries, with little consideration of whether the units are psychologically meaningful. Although not all alignment work is strictly turn-based, turns are a natural unit of aggregation in spontaneous dialogue, and the robustness of existing results to more psychologically-relevant segmentation remains unknown. Equally important, improper turn boundaries may have hidden entire classes of alignment effects that only become detectable once transcripts are partitioned in naturalistic ways. NaturalTurn tackles these issues head-on.

***Summary.*** NaturalTurn's guiding question is participant-centered: "When would the people in this conversation say one speaker's turn ended and the next began?" Its practical constraint is minimalism: given only the words and time-stamps that any off-the-shelf ASR system (i.e., transcription service) produces, can turn boundaries be segmented automatically? The algorithm (described in detail in the next section) does two basic things. First, it attaches brief listener vocalizations, such as canonical backchannels, short acknowledgements, and other feedback, to the speaker's ongoing primary turn instead of counting each as a separate turn; and second, it treats everything a speaker says as a single turn until the speaker falls silent for a parameterized, user-tunable interval that ordinarily signals floor release.

In effect, NaturalTurn uses the same raw cues available to early VAD-plus-channel-switch systems—who is talking, who is silent, and for how long—but recombines them so that the floor stays with the current speaker unless a silence long enough to matter occurs. By deliberately stopping there and eschewing prosodic features, gaze trackers, sequence models, etc., NaturalTurn remains deterministic, transparent, and scalable, able to segment millions of turns without extra annotation, training data, or feature engineering.

After detailing NaturalTurn's mechanics, we illustrate its utility on the CANDOR corpus [71], a large, publicly available set of dyadic conversations accompanied by rich post-interaction ratings of rapport, enjoyment, and impression formation.

Analyses show that NaturalTurn has the potential to advance inquiry on turn duration, gap timing, and backchannel dynamics at scale, as well as reshape alignment and synchrony findings once turns are segmented in psychologically plausible ways. These demonstrations are illustrative and not intended as comprehensive; a far wider range of turn-taking dynamics and social outcomes (i.e., trust, persuasion, conflict reduction, etc.) hinges on accurately measuring who held the floor, for how long, and amid how much parallel speech. At present, the robustness of many links between turn-taking and social outcomes is uncertain because prior work relied on coarse or inconsistent segmentation. Worse, incorrect boundaries may have obscured entire classes of effects that only emerge when transcripts are partitioned into turns that match participants' lived experience of who held the floor. By supplying that measurement in an automated, scalable, and openly shareable form, NaturalTurn lays the groundwork for a more cumulative social science of conversation.

## Methods

Our procedure to generate and analyze data was as follows: (1) select a large corpus of recorded dyadic conversations, (2) convert audio recordings into semi-structured JSON text objects; (3) apply a rudimentary turn-segmentation procedure to generate a set of baseline "stereo-separated" transcripts; (4) modify the baseline transcript segmentation using the NaturalTurn algorithm to separate primary conversational turns from secondary conversational turns, and (5) compare the baseline transcripts' descriptive and inferential properties to those generated by NaturalTurn.

**The CANDOR Corpus**

The testing ground for NaturalTurn was the CANDOR corpus, a large multimodal dataset of naturalistic conversation [71]. Released in 2023, CANDOR includes over 1TB of video-recorded conversations and survey responses from a large, diverse sample of participants in the United States. The participants were 18+ years old, recruited on the crowd-sourcing platform Prolific, and consented to have their data made public for research purposes. Unacquainted participants were paired to talk based solely upon their availability and engaged in an unstructured conversation for at least 25 minutes ($M$ = 31.3 min, $SD$ = 7.96). Data collection yielded a total of 1,656 dyadic conversations and 1456 “unique” participants (more than 50% of our sample had two or more conversations). Participants who had multiple conversations did not do so back-to-back; rather, they had several conversations distributed across the data collection period. Following each conversation, the participants completed a post-conversation survey that measured their traits, experiences, perceptions of their conversation partner, and other items. The participants received $0.85 for signing up and an additional $14.15 upon completing the conversations and the survey.

All methods complied with relevant guidelines and regulations; CANDOR’s original protocol was approved by Ethical & Independent Review Services (E&I Review; protocol #19160-01), and all participants (18+) provided informed consent for public release of their recordings and surveys.

**Audio-to-text conversion**

For each conversation in the corpus, we processed the audio recordings into transcripts using two different speech-to-text services: Amazon Web Services’ (AWS) Transcribe and Deepgram’s Nova. Two services were used rather than one because their methodologies differ and result in different raw token outputs (i.e., words). As NaturalTurn operates on these tokens to segment turns, its results necessarily depend upon the specific speech-to-text process used. Here, we report results using AWS, which had a number of features better suited to our downstream analyses, but the findings are, to our knowledge, robust with respect to different transcription services and variation in token output.

**Segmentation into baseline and NaturalTurn transcripts**

We produced a baseline transcript for each conversation by applying an elementary turn model to speech-to-text JSON output. Each conversation file consisted of a list of JSON objects, each of which represented a single spoken word (i.e., "speech token"), together with start and stop timestamps. These token objects were sorted chronologically and separated by stereo channel, with Speaker A in the left channel and Speaker B in the right. A simple turn model assigned each word token to Speaker A's turn until there was a word token from Speaker B, at which point Speaker B's turn began (Fig. 1). Although limited in many respects, this can be considered a "baseline" method to construct turns, providing a useful benchmark for improved turn segmentation algorithms. We then fed the baseline transcripts to NaturalTurn, which we describe in detail below.

**Post-processing**

Finally, we performed an analysis of the transcripts produced by the baseline algorithm, transcripts produced by the NaturalTurn algorithm, and the post-conversation survey data from all participants in the CANDOR corpus.

It should be noted that while current speech-to-text services are designed to output formats like JSON objects rather than conversational turns, certain platforms like AWS do offer interested users a tool to segment transcripts into turns. These AWS-generated turns bear a strong resemblance to what was defined here as baseline transcripts. To our knowledge, no existing speech-to-text service currently offers a feature that attempts to produce transcripts that reflect naturalistic conversational turns, which is the critical function that our algorithm addresses.

**The NaturalTurn Algorithm**

The representation of speech in alternating turns is essential to many forms of conversation analysis. NaturalTurn improves on a Baseline turn model by isolating parallel speech, defined as backchannels, simultaneous speech, and other listener vocalizations. A NaturalTurn transcript retains parallel speech content and timing, but visually and functionally separates these secondary events from primary turns, which results in more naturalistic transcripts.

NaturalTurn's full mechanics are nuanced, but its core logic is straightforward: The algorithm first works within each speaker, merging consecutive utterances separated by less than a user-set "max_pause" threshold (e.g., 1.5 s) into continuous turns. This produces each speaker's provisional turn boundaries. Then it examines overlap: any listener utterance that falls entirely within the primary speaker's turn boundaries is classified as secondary (backchannel, brief overlap, attempted interruption, etc.). Any listener utterance that starts after or extends beyond the speaker's turn boundary becomes the new primary turn.

Although NaturalTurn has more than a dozen tunable parameters, the main lever affecting turn segmentation is "max_pause." Proper calibration is critical: set too high, it creates artificially long turns that misclassify too many utterances as secondary; set too low, it fragments naturalistic speaking turns into pieces. While a more sophisticated solution might adapt this threshold to individual and contextual variation, NaturalTurn currently requires max_pause to be set at a specific value.

Using the CANDOR corpus, a max_pause of 1.5s proved effective. In contrast, increasing max_pause to 2.0 seconds or greater tended to merge turns mistakenly. Using values from 0.5s to 1.5s we observed reasonable results, but these shorter pauses tended to fragment turns. Given the specifics of the CANDOR corpus and our hypotheses related to turn dynamics, we ultimately selected a max_pause of 1.5s to strike a balance between these considerations. Overall, NaturalTurn streamlines baseline transcripts by collapsing over brief interruptions within a speaker's turn, thereby generating longer contiguous primary turns. NaturalTurn further categorizes secondary speech as either *Backchannel* or *Secondary Turn* or *Other*, a simple structure that future development may refine. For additional details, see the public release code along with baseline and NaturalTurn transcripts for every conversation in the CANDOR corpus (https://github.com/betterup/natural-turn-transcription; https://osf.io/xy35p/).

To illustrate, consider the transcript from a CANDOR conversation shown in Figure 1. The text is the opening of a conversation in which Speaker 1 is introducing himself while Speaker 2 signals understanding and engagement with brief acknowledgements such as "yeah" and "mhm." The baseline transcript (Panel A) treats every acknowledgement as a new turn, fragmenting what the likely participants experienced. In contrast, NaturalTurn treats these acknowledgements as parallel or secondary speech attached to the primary turn (Panel B). Speaker 1 keeps the floor until he remains silent for longer than the user-tunable max_pause, at which point his turn ends.

NaturalTurn further labels secondary speech as *Backchannel* if it matches a configurable cue list and satisfies three lightweight rules: (i) at most three words, (ii) does not begin with a prohibited list of starter words such as "I'm," and (iii) at least half of its tokens are on the backchannel cue list. All parameters (length limit, default filter, and cue list) are tunable so researchers can adapt them to their own corpora and research questions. Approximately two-thirds of all the secondary speech isolated by NaturalTurn was labeled as *Backchannel*.

When listener contributions are more substantial, these are labeled *Secondary Turns*. Figure 1, Panel B shows an *Intermediate* transcript (what remains after backchannels are stripped out) of a storytelling stretch in which the listener interjects phrases such as "Oh my God" and "Just wait for them." Even though they do not meet the criteria for backchannels, NaturalTurn separates these interjections and keeps the entire story as one primary turn. Both *Backchannel* and *Secondary Turn* content are stored separately from the speaker's primary contribution, with timing and content retained for further analysis.

---------------------------------

**Figure 1**

*Turn Segmentation Using Baseline and NaturalTurn Models.*

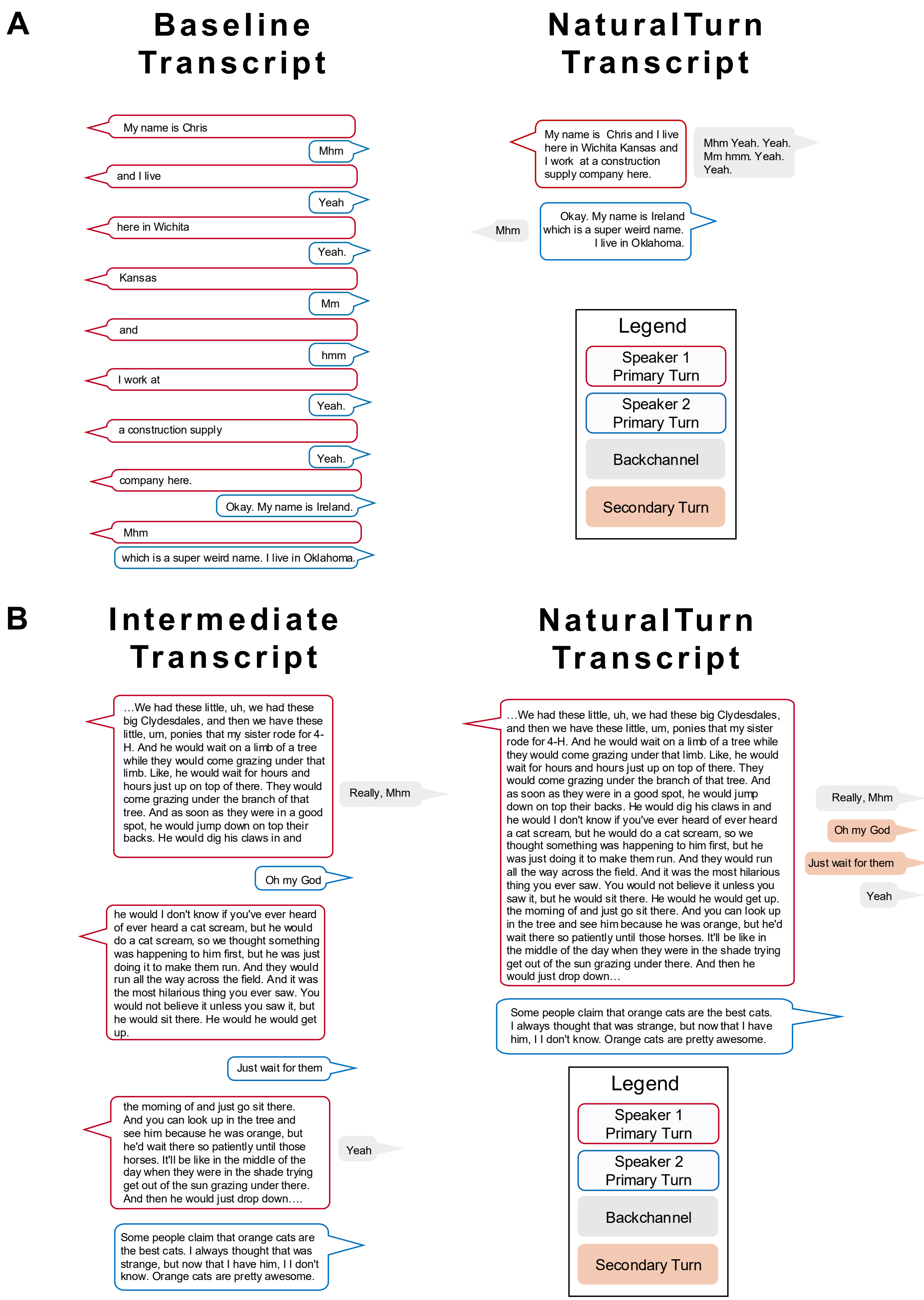
A
Baseline Transcript
My name is Chris
Mhm
and I live
Yeah
here in Wichita
Yeah.
Kansas
Mm
and
hmm
I work at
Yeah.
a construction supply
Yeah.
company here.
Okay. My name is Ireland.
Mhm
which is a super weird name. I live in Oklahoma.
NaturalTurn Transcript
My name is Chris and I live here in Wichita Kansas and I work at a construction supply company here.
Mhm Yeah. Yeah. Mm hmm. Yeah. Yeah.
Mhm
Okay. My name is Ireland which is a super weird name. I live in Oklahoma.
Legend
Speaker 1 Primary Turn
Speaker 2 Primary Turn
Backchannel
Secondary Turn
B
Intermediate Transcript
…We had these little, uh, we had these big Clydesdales, and then we have these little, um, ponies that my sister rode for 4-H. And he would wait on a limb of a tree while they would come grazing under that limb. Like, he would wait for hours and hours just up on top of there. They would come grazing under the branch of that tree. And as soon as they were in a good spot, he would jump down on top their backs. He would dig his claws in and
Really, Mhm
Oh my God
he would I don't know if you've ever heard of ever heard a cat scream, but he would do a cat scream, so we thought something was happening to him first, but he was just doing it to make them run. And they would run all the way across the field. And it was the most hilarious thing you ever saw. You would not believe it unless you saw it, but he would sit there. He would he would get up.
Just wait for them
the morning of and just go sit there. And you can look up in the tree and see him because he was orange, but he'd wait there so patiently until those horses. It'll be like in the middle of the day when they were in the shade trying get out of the sun grazing under there. And then he would just drop down….
Yeah
Some people claim that orange cats are the best cats. I always thought that was strange, but now that I have him, I I don't know. Orange cats are pretty awesome.
NaturalTurn Transcript
…We had these little, uh, we had these big Clydesdales, and then we have these little, um, ponies that my sister rode for 4-H. And he would wait on a limb of a tree while they would come grazing under that limb. Like, he would wait for hours and hours just up on top of there. They would come grazing under the branch of that tree. And as soon as they were in a good spot, he would jump down on top their backs. He would dig his claws in and he would I don't know if you've ever heard of ever heard a cat scream, but he would do a cat scream, so we thought something was happening to him first, but he was just doing it to make them run. And they would run all the way across the field. And it was the most hilarious thing you ever saw. You would not believe it unless you saw it, but he would sit there. He would he would get up. the morning of and just go sit there. And you can look up in the tree and see him because he was orange, but he'd wait there so patiently until those horses. It'll be like in the middle of the day when they were in the shade trying get out of the sun grazing under there. And then he would just drop down…
Really, Mhm
Oh my God
Just wait for them
Yeah
Some people claim that orange cats are the best cats. I always thought that was strange, but now that I have him, I I don't know. Orange cats are pretty awesome.
Legend
Speaker 1 Primary Turn
Speaker 2 Primary Turn
Backchannel
Secondary Turn

*Note.* A) Audio-to-text segmentation using the baseline turn model treats each listener interjection as a new turn, whereas NaturalTurn isolates listener backchannels, leaving only speakers' alternating introductions. B) Even with backchannels removed, parallel speech still remains, breaking up a speaker's story artificially. NaturalTurn addresses this by removing these routine interjections, leaving the speaker's story fully intact. (CANDOR corpus conversation ID 00411458-8275-4b92-a000-d52187f03604)

---------------------------------

For maximum research utility, NaturalTurn has multiple tunable parameters. When adapting NaturalTurn to a new dataset there are five main decision-points that researchers should consider. For each, we flag the key user-facing parameter(s), along with bracketed items that represent secondary options. Together these five parameter groups capture the practical choices a user makes when adapting the algorithm. Full information is available in the public code release.

1. Collapse across short pauses
*max_pause*

Consecutive utterances from the same speaker are stitched together whenever the silence gap between them is ≤ max_pause. During that full stretch of stitched together utterances, any words spoken by the *other* channel (i.e., the current listener) are tagged as non-primary turns so they cannot break up the flow of the primary turn.
[collapse_short_pauses (on/off flag)]

2. Identify listener backchannels
*backchannel_cues, backchannel_word_max*

Non-primary utterances that match a backchannel_cues list (e.g., mhm, yeah) and fall below the word-count cutoff (*backchannel_word_max*) are labelled as backchannels.
[ backchannel_proportion, not_backchannel_cues, backchannel_second_max, backchannel_pause_max]

3. Classify secondary turn versus other
*not_backchannel_cues, backchannel_word_max*

A non-primary utterance is labeled "secondary" when it either (i) exceeds the short-feedback length cutoff ( n_words > backchannel_word_max ) *or* (ii) begins with a short clause-opening token in not_backchannel_cues (e.g., “and”, “but”, “well”). All remaining short listener fragments fall into the “other” bucket. [backchannel_second_max, backchannel_proportion ]

4. Handle low-confidence and problem tokens
*token_confidence_threshold*

Words whose ASR recognition probability drops below the threshold can be (a) removed, (b) wrapped in brackets, or (c) replaced with <<unknown>>. [exclude_underconfident_utterances, flag_underconfident_tokens, mask_underconfident_tokens]

Strip problem tokens
*banned_tokens, suspicious_tokens*

Anything in banned_tokens is always removed (e.g., known idiosyncratic mis-recognitions); items in suspicious_tokens are removed only when they are also low-confidence. [remove_banned_tokens, remove_suspicious_tokens]

5. Choose the output shape
*combine_secondary_turns, pivot_secondary_turns*

Decide whether secondary material is (i) nested inside the same data-frame row as the primary turns, and/or (ii) separated into “speaker” versus “listener” columns.

The category labels produced by NaturalTurn are pragmatic heuristics, not definitive functional classes. The default rules that funnel fragments into *Backchannel*, *Secondary*, or *Other* offer a sensible starting point, but they will seldom align perfectly with every theoretical taxonomy. In practice, many items marked secondary could just as plausibly be treated as extended backchannels. A concrete illustration comes from the transcripts analyzed in this paper. During our study, an automatic update to our speech-to-text service began spelling “okay” as “ok.” Because the new token was missing from our cue list, many genuine backchannels were mis-routed into the *Other* bucket. Although this event does not alter the substantive findings reported here, it is a cautionary tale: routine auditing of cue lists, length cut-offs, and confidence thresholds, with corpus-specific adjustments as needed, is essential for accurate deployment.

In summary, the goal of NaturalTurn is to segment conversational transcripts the way that participants themselves would recognize. Our algorithm improves on baseline transcripts which currently lack the means to address parallel speech, resulting in turns that are psychologically inconsistent with the natural flow of conversation. The tunable features of this algorithm make it adaptable to a variety of practical problems and research questions. In the next section, we verify NaturalTurn's ability to more accurately represent basic turn-taking dynamics and then link these dynamics to features of conversations which indicate successful social interactions.

**Using NaturalTurn to Analyze Turn-taking Dynamics and Psychological Outcomes**

The turn-taking dynamics of two people talking contain a wealth of social information. But if we lack the means to format transcripts into naturalistic turns at scale, most of these dynamics will remain empirically and experimentally underexplored. In this regard, NaturalTurn represents a step forward as it better approximates two fundamental characteristics of conversational turns: how long each speaker talks (durations) and gaps over overlaps between speakers (intervals). As shown in Figure 2, NaturalTurn has a considerable effect on the sequencing and measurement of these components, underscoring the critical role of turn segmentation algorithms in the science of social interaction.

----------------------------------

**Figure 2**

*NaturalTurn's Effect on Turn-taking Dynamics.*

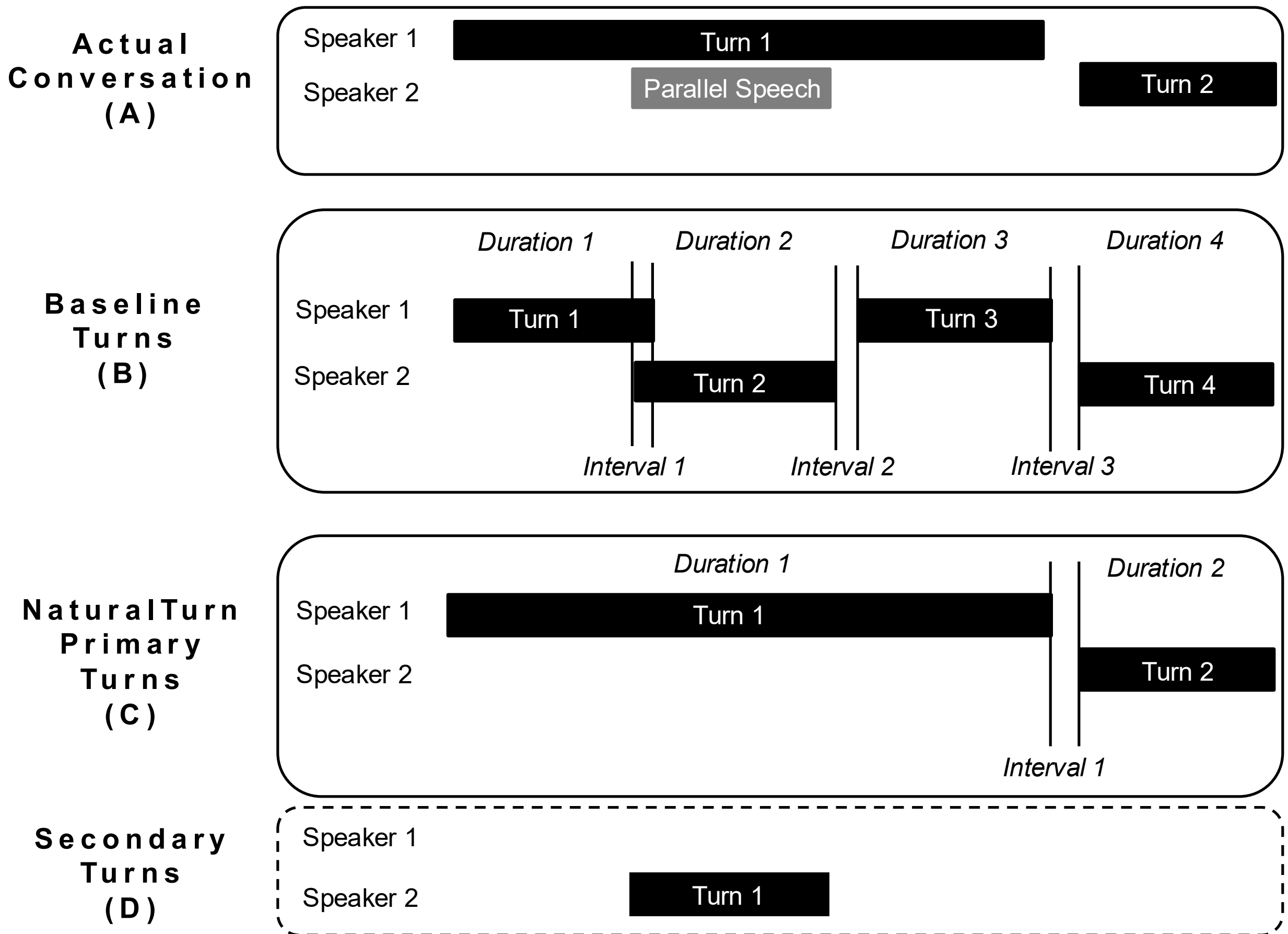

*Note.* NaturalTurn has a major influence on the sequencing and measurement of conversational turns. When a conversation contains parallel speech (depicted in A as an overlap between Speaker 1 and Speaker 2), baseline and NaturalTurn models diverge considerably (compare panels B and C/D), resulting in different values for turn durations and the intervals between turns, two core features of the turn-taking system. Specifically, compare the baseline model's series of short overlapping turns (baseline Turns 1-3) to the NaturalTurn model's single long turn (NaturalTurn Turn 1). Further, what is recorded as three intervals between the four Baseline turns, consisting of both gaps and overlaps, becomes just one interval between the two NaturalTurn turns, a single gap. NaturalTurn's transformations have a considerable effect on the empirical dynamics of turn-taking, underscoring the critical role of turn segmentation algorithms in building a robust science of social interaction.

----------------------------------

To prepare the transcripts, we removed the first turn from each conversation, as conversations in the CANDOR dataset sometimes begin with one person waiting for the other to join the video chat. During that waiting period, speech from off-camera conversations and other background noise are occasionally (and mistakenly) included as a transcript's first "turn".

Before summary statistics were calculated, we identified and removed exceptionally long speaking turns and intervals between turns, which were frequently caused by technical issues with the call, such as poor Internet connectivity or audio echo. Specifically, we set to "NA" (i) turns longer than 120 s (<0.5 % of turns); and (ii) intervals between turns whose absolute value exceeded 5 s (<1 % of turns). Note that a positive interval represents a gap between consecutive turns, whereas a negative interval represents an overlap. Sensitivity analyses showed that removing these turns did not affect our overall conclusions.

Aggregations were then performed at the turn level, producing summary measures for each turn: mean turn duration, mean words per turn, mean turn interval, and proportion of negative turn intervals (i.e., overlapping turns). Aggregation at the speaker level produced another metric, mean number of turns per speaker per conversation. Mean turn duration was then associated with the three post-conversation survey responses, in which the participants evaluated their experience of the conversation and their partner. *Enjoyment* was a single-item rating ("How enjoyable was the conversation?", 1 = Not at all, 9 = Extremely); *Affect* was a single valence item ("How positive or negative do you feel right now?", 1 = Extremely negative, 9 = Extremely positive); and *Shared Reality* was the 8-item Shared Reality Scale ([72]; e.g., "We developed a joint perspective," 1 = Strongly disagree, 7 = Strongly agree). This allowed us to explore the relation between granular turn dynamics and speakers' post-conversation impressions.

***Statistical Analyses.*** We ran all statistics in R. Simple associations between a speaker's mean turn duration and the three post-conversation ratings (enjoyment, overall affect, shared reality) were evaluated with Pearson correlations. Because each dyad contributes two speakers nested in a conversation, we checked robustness with a linear mixed-effects model that included a random intercept for conversation ID. We extracted standardized effect sizes from the mixed models and converted regression coefficients back to correlation coefficients. Results from the multilevel model were virtually identical to the Pearson correlations, so we report the simpler correlations in the main text and figure.

## Results

Comparisons of central tendency (Table 1a) and distributional characteristics (Figures 3 and 4) highlight the differences between the two models. Notably, the mean turn duration in NaturalTurn transcripts was more than 400% longer than in baseline transcripts (12.6s vs 2.9s, $p < 0.001$), which is more representative of the conversations' natural flow. Considering the range of turn durations, NaturalTurn transcripts returned a broad distribution (0-20+ seconds; see Table 1 in [73], for a comparison). In stark contrast, most baseline turns were artificially brief, typically under 2 seconds, which is not an accurate representation of human dialogue. Similarly, NaturalTurn transcripts had more words per turn and significantly fewer turns per conversation overall. The median statistics (Table 1b) show even larger differences between the two approaches.

Consider that analyses of large-scale corpora typically report mean inter-pausal-unit (IPU) lengths of about 2 s [9]. Because a psychologically meaningful turn normally bundles several such units, natural conversational turns should be longer. Yet our baseline model yields an average turn duration of only 2.9 s and the majority below 2 s, approximating a single IPU. In other words, baseline turns are "IPU-sized," not turn-sized, revealing how the model fragments the underlying dialogue. By removing parallel speech and re-merging consecutive IPUs from the same floor-holder, NaturalTurn recovers more naturalistic durations (mean ≈ 12 s), providing units that better align with conversational intuition and prior literature.

NaturalTurn's output also alters the transcript's representation of how people exchange turns. As shown in Table 1a, 60% of baseline turns had at least some overlap with the preceding turn, meaning that the new speaker begins before the previous speaker has finished. By contrast, only 35% of NaturalTurn turns had overlaps. Further, as shown in Table 1a and Figure 4, the mean time for the exchange of turns was significantly greater for NaturalTurn transcripts (+171 ms) compared to baseline (+39 ms) ($p < 0.001$). While this difference may seem small, these milliseconds are pivotal, as precise timing acts as a significant cognitive constraint on the potential mechanisms that underlie human turn exchange. Indeed, NaturalTurn turn intervals are squarely within the 100-200 ms estimate frequently proposed in the literature [9, 56, 74]. These results reinforce how one's choice of turn-segmentation algorithm shapes "the ground truth" (i.e., transcript structure) upon which researchers must rely.

Finally, although the present paper focuses on primary turns and their links to social outcomes, NaturalTurn's secondary speech classifier is itself a valuable object of study. Although large-corpus benchmarks are sparse, our descriptive statistics sit comfortably within existing ranges. For example, across the CANDOR conversations we analyzed, listener backchannels occurred ≈ 3.5 times per minute, matching reported rates for vocal backchannels [75]. Moreover, the mean duration of a turn labelled *Backchannel* was ≈ 750 ms, which is similar to estimates in the literature [76]. *Secondary turns* were, on average, longer ( ≈ 1.5 s), reinforcing the idea that the algorithm is reserving the *Backchannel* label for briefer instances of feedback while treating extended listener feedback as a distinct class of parallel speech. These findings suggest that the algorithm's automatic labels recover known phenomena previously identified by careful annotation, and thus may facilitate large-scale labeling and investigation of parallel speech dynamics.

---------------------------------

**Table 1a**

*Comparison of Baseline and NaturalTurn Models: Means*

| Statistic | Baseline Mean | BL SD | BL 95 % CI | NaturalTurn Mean | NT SD | NT 95 % CI | Test | *p* |
|---|---|---|---|---|---|---|---|---|
| Turn duration (s) | 2.90 | 5.26 | [2.89 – 2.91] | 12.63 | 16.13 | [12.56 – 12.70] | t = 284.5 (df = 238348) | < .001 |
| Words per turn | 8.80 | 16.13 | [8.76 – 8.83] | 36.53 | 48.59 | [36.33 – 36.73] | t = 269.1 (df = 238736) | < .001 |
| Interval between turns (ms) | 38.65 | 522.13 | [37.65 – 39.65] | 170.99 | 1174.53 | [166.15 – 175.84] | t = 52.4 (df = 245110) | < .001 |
| Proportion negative intervals (overlap) | 0.62 | 0.49 | [0.62 – 0.62] | 0.35 | 0.48 | [0.35 – 0.35] | $\chi^2$ = 54624.8 (df = 1) | < .001 |
| Turns per speaker per conversation | 314.99 | 130.46 | [310.54 – 319.43] | 69.05 | 27.08 | [68.12 – 69.97] | t = 115.9 (df = 3311) | < .001 |

**Table 1b**

*Comparison of Baseline and NaturalTurn Models: Medians*

| Statistic | Baseline Median | BL IQR range | NaturalTurn Median | NT IQR range |
|---|---|---|---|---|
| Turn duration (s) | 0.74 | [0.41 – 3.25] | 6.58 | [2.79 – 15.68] |
| Words per turn | 2.00 | [1.00 – 10.00] | 18.00 | [8.00 – 45.00] |
| Interval between turns (ms) | -71.00 | [-259.00 – 201.00] | 320.00 | [-389.00 – 809.00] |

*Note.* Both mean (Table 1a) and median (Table 1b) statistics reveal consistent patterns: NaturalTurn produces substantially longer turns with more words compared to the baseline model. Specifically, NaturalTurn turns are approximately 9× longer by median (6.58 vs 0.74 s) and 4× longer by mean, containing 9× more words by median and 4× by mean. This results in conversations with roughly one-quarter as many total turns per speaker (69 vs 315).

Pronounced differences also appear in turn interval distributions. For the baseline model, the median interval is negative (-71 ms, indicating overlap) while the mean is positive (39 ms), revealing a distribution dominated by overlaps. NaturalTurn shows larger intervals overall, with positive values for both median (320 ms) and mean (171 ms). The proportion of overlapping intervals is nearly halved under NaturalTurn (0.35 vs 0.62).

For continuous measures, we report means ± SD with 95% confidence intervals in Table 1a, and medians with interquartile ranges [Q1–Q3] in Table 1b. All statistics except the final row were calculated across individual turns; the final row aggregates at the speaker level. Welch's *t*-tests compared means, a $\chi^2$ test compared overlap proportions, and paired *t*-test for the speaker-level comparison. For bounded distributions (durations, word counts), SDs indicate spread rather than implying negative values. Similarly, the SD for overlap proportions reflects binomial variance and should not be interpreted as a range.

---------------------------------

**Figure 3**

*Distribution of Turn Durations for baseline and NaturalTurn Models*

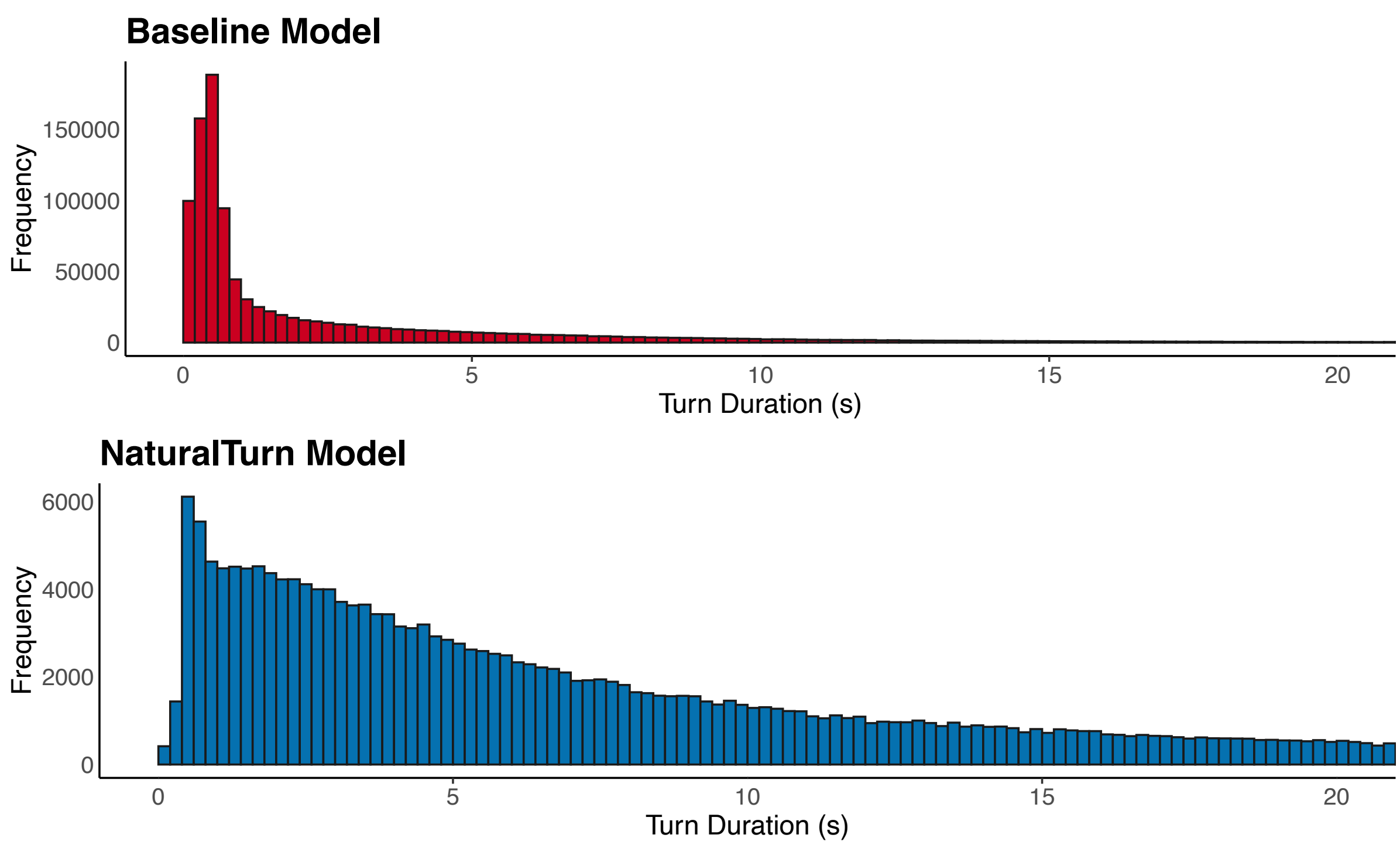

*Note.* NaturalTurn's effect on turn duration was significant and of considerable magnitude. Baseline turns (red) are short and fragmented (median = 0.74 s), whereas NaturalTurn turns (blue) are longer (median = 6.58 s). The latter are more naturalistic, and better suited for many research questions.

---------------------------------

---------------------------------

**Figure 4**

*Distribution of Turn Intervals for Baseline and NaturalTurn Models*

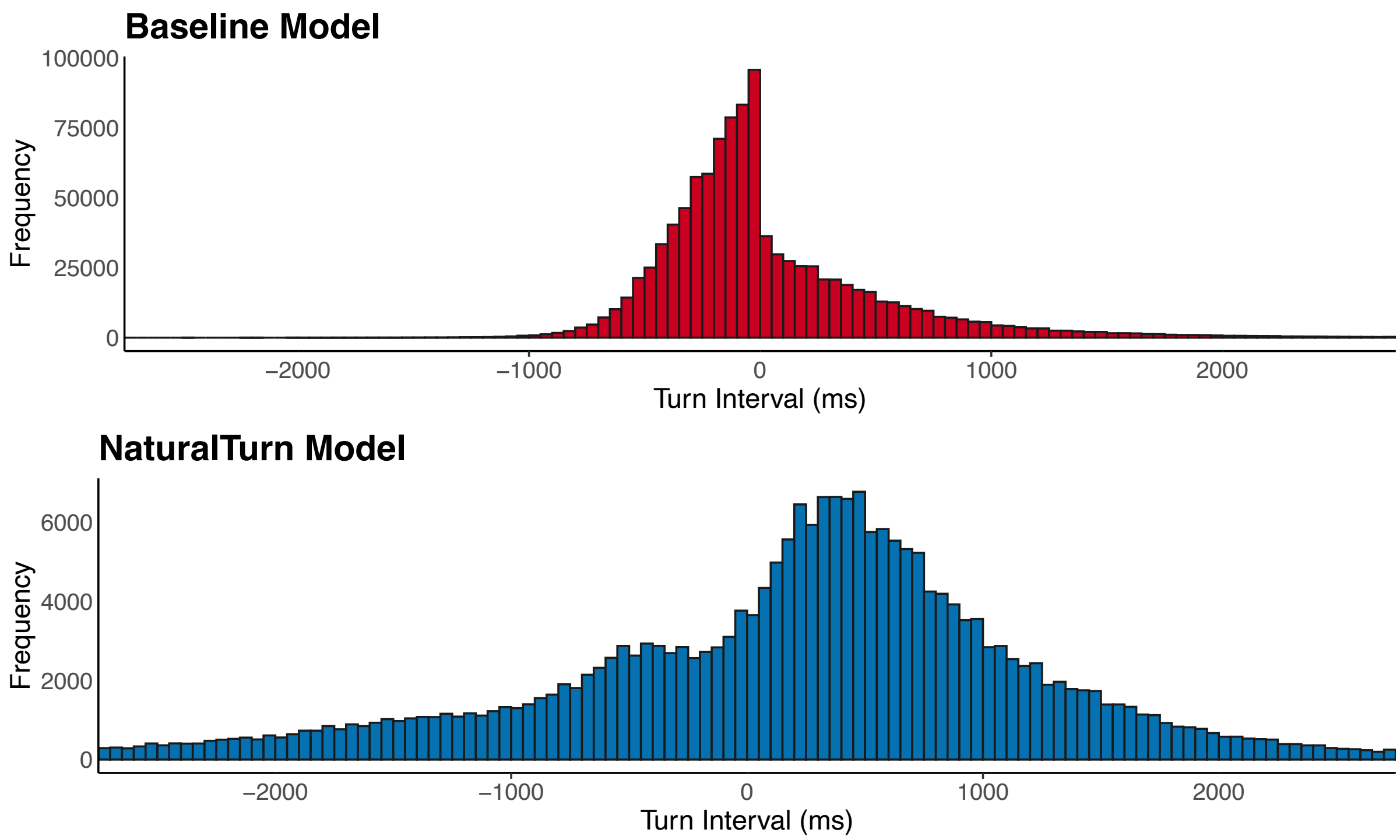

*Note.* Baseline turn intervals (red) are predominantly overlaps (negative values) with a negative median ( -71 ms). In contrast, the NaturalTurn model (blue) shows a much wider distribution with the median shifted to the right (320 ms) indicative of more conversational gaps. This brief gap between turns matches both existing literature and everyday experience of natural conversation flow.

---------------------------------

## Turn Duration and Enjoyment

Figure 5 illustrates a notable correlation between speakers' mean turn duration and their reported enjoyment of the conversation—but only for NaturalTurn-generated turns. For Baseline turns, the correlation between mean turn duration and enjoyment was roughly zero ($r$ = 0.002 , 95% CI = [-0.03, 0.04], $t(3255) = 0.14$, $p = .89$) By contrast, for NaturalTurn turns, the relation between mean turn duration and enjoyment was positive and significant ($r$ =0.14, 95% CI = [0.10, 0.17], $t(3255) = 7.90$, $p < .001$. A test of the difference between two paired correlations revealed that the correlation produced by our NaturalTurn model was significantly greater than that of the Baseline turn model ($t(3252) = 8.51$, $p < .001$). Clustering standard errors by conversation using a multilevel model yielded similar results. Although the magnitude of this correlation is small, the NaturalTurn model reveals an intuitively quite plausible link between longer turn durations and increased conversational enjoyment that is not discernible with Baseline turns.

Complementing the enjoyment results, we also observed significant positive correlations between speakers' mean NaturalTurn turn durations and two additional important post-conversation measures: overall affect ($r = 0.13$, 95% CI = [0.09, 0.16], $t(3255) = 7.35$, $p < .001$), and one's sense of shared reality with one's conversation partner ($r = 0.13$, 95% CI = [0.09, 0.16], $t(3255) = 7.22$, $p < .001$). No significant correlations were observed when using baseline turns to predict these outcomes (affect: $r = -.01$, 95% CI = [-0.04, 0.03], $t(3255) = -0.50$, $p = .61$; shared reality: $r = -.03$, 95% CI = [-0.07, 0.00], $t(3255) = -1.90$, $p = .06$). Linking turn-taking dynamics to participants' post-conversation affective reactions appears to require the ability to segment transcripts into naturalistic turns.

Although NaturalTurn improves the structure and flow of conversational transcripts, its utility as an input into statistical modeling will not extend uniformly across all research areas. The correlations we report, while statistically significant and theoretically consistent, involve just one low-level feature (mean turn duration) and a small set of subjective outcomes. Future studies will need to explore how segmentation decisions affect downstream modeling in other domains, contexts, or populations. Our aim here is not to claim that any specific effect size is definitive, but rather to illustrate how analytic conclusions can shift depending on the turn model used.

---------------------------------

**Figure 5**

*Turn Duration and Enjoyment Across Turn Models*

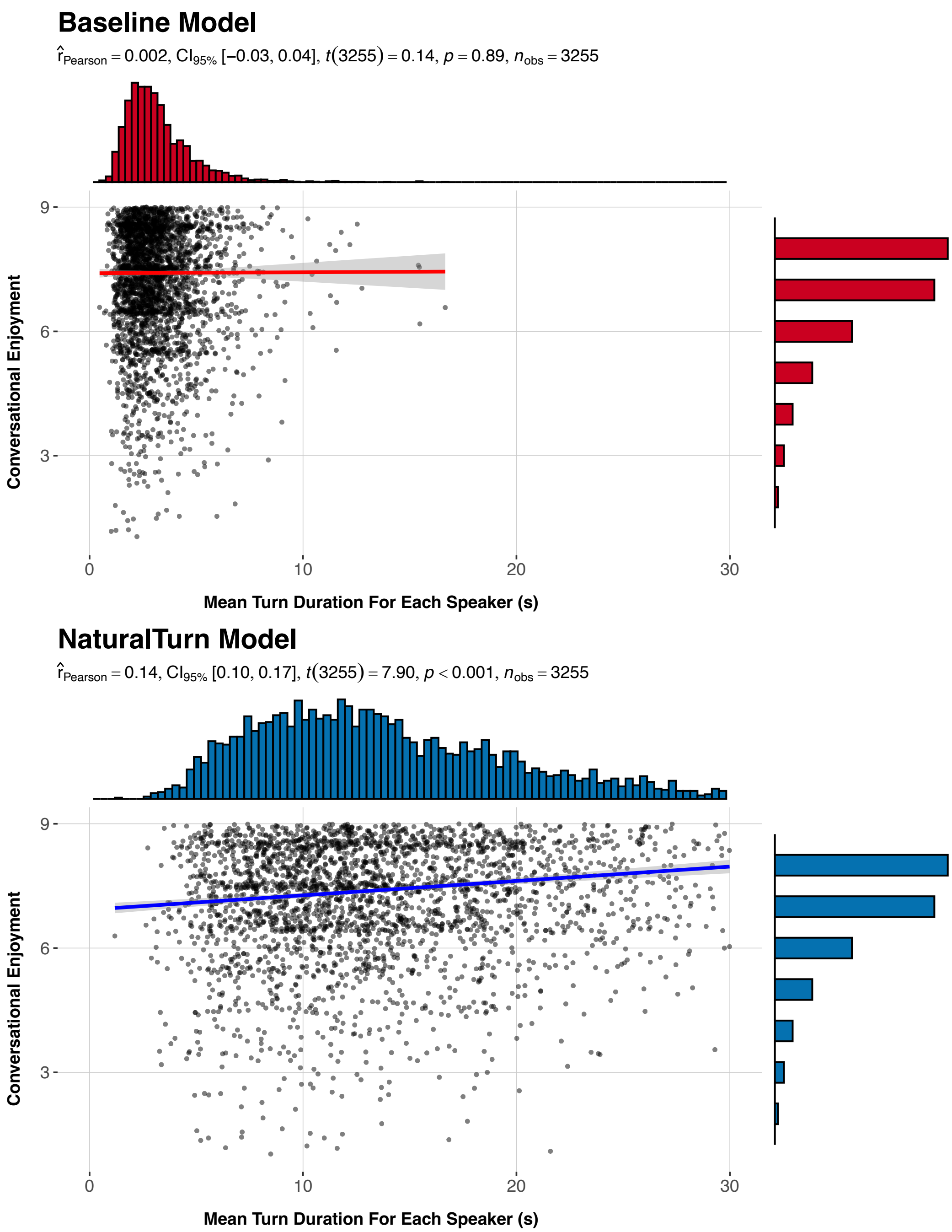

*Note.* Speakers' mean turn duration is related to their reported conversational enjoyment—but only for NaturalTurn-generated turns. The x-axis represents participants' mean turn duration averaged across each conversation. The y-axis represents their reported enjoyment after their conversation ended. While longer duration turns were positively associated with enjoyment for NaturalTurn turns, no such relation was found for the turns in the baseline transcripts. This underscores the challenges of connecting lower-level

features of conversation with important higher-level conversational outcomes, together with the methodical and algorithmic advances necessary to do so.

---------------------------------

**Conversational Alignment**

To further demonstrate NaturalTurn's utility, we examined how turn segmentation methods affect the measurement of conversational alignment, a set of phenomena whereby dialogue partners coordinate their linguistic behaviors during interaction. This coordination manifests in multiple ways: speakers may synchronize by matching each other's moment-to-moment behavioral patterns, entrain by adapting their timing and rhythms, or accommodate by adjusting their language style to match their partner. While researchers use various terms for these phenomena (convergence, synchrony, entrainment, accommodation, adaptation, mirroring, etc.), all capture the fundamental concept of coordinated, co-dependent behavior. Alignment has been linked to enhanced rapport, mutual understanding, and relationship quality [62-69]. Accurate turn segmentation is particularly critical for alignment research: when transcripts are fragmented by artificial turn boundaries or overlapping speech, measures of how speakers coordinate—whether converging, synchronizing, or otherwise aligning—may be systematically distorted.

***Method.*** We analyzed the same conversations from the CANDOR corpus, measuring alignment across three dimensions: Language Concreteness, Positive Affect, and Arousal/Activation. These dimensions were selected to capture distinct aspects of linguistic behavior (semantic content, emotional valence, and physiological activation) that are fundamental to interpersonal coordination.

Before analysis, transcripts underwent standard preprocessing using methods derived from the ConversationAlign package [77, 78]. This included removing stopwords using a customized list that excludes common greetings, idioms, filler words, numerals, and pronouns. Text was then lemmatized to transform inflected forms (e.g., “standing,” “stands”) into their root forms (e.g., “stand”), enabling accurate quantification.

Each dimension was then quantified using validated psycholinguistic dictionaries. Language Concreteness reflects the degree to which words refer to perceptible, tangible concepts versus abstract ideas, with values derived from crowd-sourced ratings of 40,000 English words [79]. Concrete words like “table” or “dog” receive high scores, while abstract words like “justice” or “theory” receive low scores. Positive Affect captures emotional positivity in language, measured using AffectVec [80], which uses vector-based embeddings to calculate semantic similarity between each word and emotional anchor words, in this case, “happiness.” Arousal/Activation measures the extent to which words cause physiological or psychological activation, based on crowd-sourced ratings of 20,000 words from the NRC-VAD lexicon [81]. All values were rescaled to a 0-10 range for comparability (see ConversationAlign documentation for full preprocessing details [77, 78]).

Briefly, each word in the cleaned and lemmatized transcripts was assigned a numeric score for the relevant dimension. Word-level scores were then averaged within each speaker’s turn to yield turn-level scores. As there are two speakers per conversation, this leads to two time-series of scores for each of the three dimensions per conversation (e.g., Positive Affect $_{\text{Speaker A}}$ vs. Positive Affect$_{\text{Speaker B}}$) which we analyzed to assess linguistic mirroring. We next formed non-overlapping “exchanges” by pairing adjacent turns (Turn 1 by Speaker A + Turn 2 by Speaker B = Exchange 1; Turn 3 + Turn 4 = Exchange 2; …) and kept each exchange’s original index so that Exchange 10, for example, always refers to the same chronological position across conversations. If every token in either turn (or in both turns) lacked a dictionary value for a given dimension, that entire exchange was flagged as “missing” for that dimension and was dropped from analysis for that conversation. Because NaturalTurn removes secondary speech turns, baseline segmentation produced many short turns that consisted entirely of backchannels (e.g. “uh-huh”, “yeah”) that rarely appear in the lexical norms. Thus, the baseline condition naturally lost more exchanges than NaturalTurn. Critically, removing these turns does not appear to inflate alignment in NaturalTurn; if anything, it makes baseline more similar to Natural-Turn and therefore constitutes a conservative test of our hypotheses. Finally, conversations that ended with an odd number of turns created incomplete final exchanges containing only one speaker; these final exchanges were removed because alignment metrics require contributions from both interlocutors. The remaining exchanges were analyzed by computing the absolute difference between speakers on each dimension, and these difference-sequences formed the basis for all alignment statistics.

It is worth noting that while our primary analyses used unigram-based methods for simplicity and interpretability, we checked our findings on a random subset of 500 conversations using BERT-based sentence embeddings projected onto axes for each dimension. This approach yielded similar patterns, suggesting that our results are robust to text representation method; and, if anything, becomes more pronounced with a contextual, sentence-level embedding.

Following an established taxonomy [70], we computed four complementary metrics of alignment:

First, *Static Local Proximity* was measured by calculating the absolute difference between speakers for each exchange, then computing the area under the curve (dAUC) across the conversation. Lower dAUC values indicate that speakers maintained closer proximity in their linguistic features throughout the conversation. To account for conversations of different lengths, we normalized dAUC by dividing by the number of exchanges. It is worth noting that this dAUC bears no relation to the "AUC" (area-under-the-ROC-curve) metric used to evaluate machine-learning classifiers.

Second, *Dynamic Local Convergence* assessed indicates whether speakers became more similar over time, assessed here by fitting quadratic regression models to the absolute-difference time series. A negative quadratic coefficient indicates a U-shaped pattern: Speakers start and end with larger differences but converge in the middle of the conversation, perhaps as they build rapport and find common ground. A positive quadratic coefficient indicates an inverted-U pattern: Speakers start and end more similar (possibly due to formulaic greetings and closings) but diverge in the middle, perhaps as they explore different topics, pursue linguistic autonomy, or express individual perspectives.

Third, *Static Global Synchrony* measures how closely the two speakers' linguistic patterns mirror each other throughout the conversation. We calculated Spearman rank correlations (ρ) between the two speakers' complete time series for each dimension. This correlation captures whether the speakers' values tend to rise and fall together on a turn-by-turn basis. Unlike parametric correlations that require linear relationships, Spearman correlation captures any monotonic association, detecting synchrony whenever speakers consistently move in the same direction, regardless of the rate of change (e.g., one speaker might show larger swings than the other, but they still move in tandem). Higher correlations indicate stronger synchrony.

Fourth, *Dynamic Global Synchrony* detects lagged dependencies in the dyadic conversation by using Pearson correlations between speakers' time series calculated at offsets ranging from -5 to +5 exchanges. A peak at lag 0 indicates tight moment-to-moment synchrony, whereas peaks at positive or negative lags suggest leader-follower dynamics. We focused on lag-0 correlations as speakers in the CANDOR corpus were randomly assigned without further categorization and thus should not systematically differ in leader-follower roles. All analyses were performed separately for baseline and NaturalTurn transcripts using identical procedures, allowing direct comparison of how turn segmentation affects alignment.

As shown in Figure 6 (left and middle panels), different turn-segmentation methods produced substantially different alignment measurements across all three dimensions. For *Static Local Proximity*, NaturalTurn transcripts showed consistently smaller differences between speakers compared to baseline (lower dAUC values; all $p$s < .001). The differences in overall normalized dAUC were substantial, ranging from 20-25%, indicating that the choice of segmentation method fundamentally alters the measured magnitude of linguistic similarity between speakers.

*Dynamic Local Convergence* patterns revealed even more striking differences between methods (Figure 6, right panels). While baseline often produced flat or even downward-curving relationships, NaturalTurn data followed U-shaped curves with significantly more negative quadratic coefficients across all dimensions (all $p$s < .001). These contrasting patterns suggest entirely different interpretations: the baseline data argue that speakers maintain relatively stable differences, while NaturalTurn suggests that speakers initially diverge, progressively converge as the conversation develops, and then end by diverging again; an intuitively plausible but otherwise hidden pattern. The same conversations thus yield very different conclusions about conversational alignment depending on segmentation method.

Global synchrony measures (Figure 7) were also different for NaturalTurn segmentation compared to baseline. Spearman correlations between speakers increased significantly for Language Concreteness, Positive Affect, and Arousal/Activation (all *p*s < .001). Notably, lag-0 correlations also showed more pronounced peaks in NaturalTurn transcripts (all *p*s < .001), where NaturalTurn's lag profiles exhibited a sharp peak at zero lag that was absent or attenuated in baseline transcripts. Turn segmentation alters interpretations of temporal dynamics; whether speakers adapt within the same exchange or with delays of several turns.

Once again, it is important to point out that NaturalTurn's differences are not uniform across all linguistic dimensions. The impact of turn segmentation decisions depends on the specific features being studied, and researchers must carefully consider how their turn segmentation choices interact with the particular dimensions of conversation and language they aim to study.

Importantly, these differences observed cannot be attributed to differences in lexical content between the two methods. Although NaturalTurn excludes low-confidence tokens during preprocessing, the median token overlap between methods was 96% (IQR = 94–97%). Moreover, when we repeated all analyses using only the 200 conversations with the lowest lexical overlap, the NaturalTurn differences remain substantively the same, confirming that changes in alignment metrics stem from differences in turn segmentation rather than token filtering.

Overall, these findings demonstrate that turn-segmentation methods strongly shape both the magnitude of alignment signals detected and the psychological patterns inferred from those signals. The same conversational data can yield evidence of either convergence or divergence, tight synchrony or loose coupling, depending on how turns are defined. While we cannot definitively claim which method is "correct," as this likely depends on the specific research question, the substantial differences observed suggest that segmentation decisions are not neutral. By isolating primary speaking turns from parallel speech, NaturalTurn (and other yet to be developed turn-models) will likely provide a cleaner starting point for detecting the nuanced ways speakers attune to one another. This has immediate implications for the large body of research linking alignment to social outcomes: without accurate turn segmentation, researchers risk underestimating, mischaracterizing, or missing entirely the conversational coordination that underlies successful human interaction.

---------------------------------

**Figure 6**

*Convergence & Turn Segmentation*

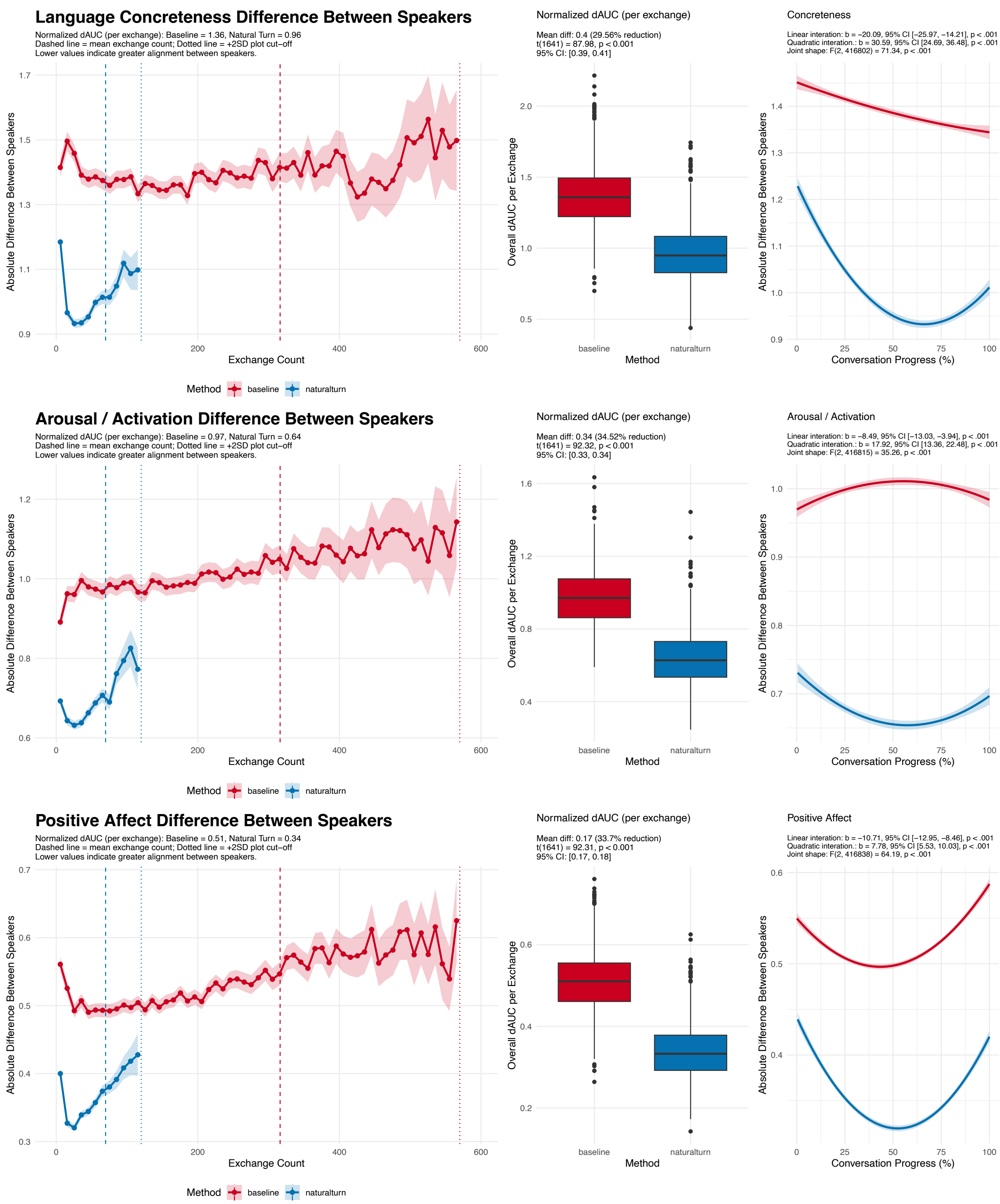

*Note.* Alignment patterns for three linguistic dimensions (Language Concreteness, Arousal/Activation, Positive Affect). Red represents baseline segmentation; blue represents NaturalTurn segmentation. Left panels: Bin-averaged absolute differences between partners' scores across the conversation (10-exchange bins; each exchange comprises two adjacent turns, one from each partner; shaded ribbons = 95%

confidence intervals). Lower values indicate stronger alignment (i.e., similarity between turns) and vertical dashed lines mark mean exchange counts for each method. Middle panels: Conversation-level alignment measured as the mean absolute difference between speakers per exchange. Boxplots show that NaturalTurn consistently reduces mean dAUC relative to Baseline, indicating speakers are more similar overall. Right panels: To test whether alignment changes over conversation progress, we regressed exchange-wise differences on relative position (0% = start, 100% = end) using orthogonal linear and quadratic contrasts. NaturalTurn exhibits steeper negative linear trends and stronger concave-up quadratic patterns, indicating increasing convergence over time that then stabilizes or diverges later, whereas Baseline shows flatter trajectories. Together, these results demonstrate that naturalistic turn segmentation yields both stronger overall alignment and more pronounced convergence dynamics.

----------------------------------

----------------------------------

**Figure 7**

*Synchrony & Turn Segmentation*

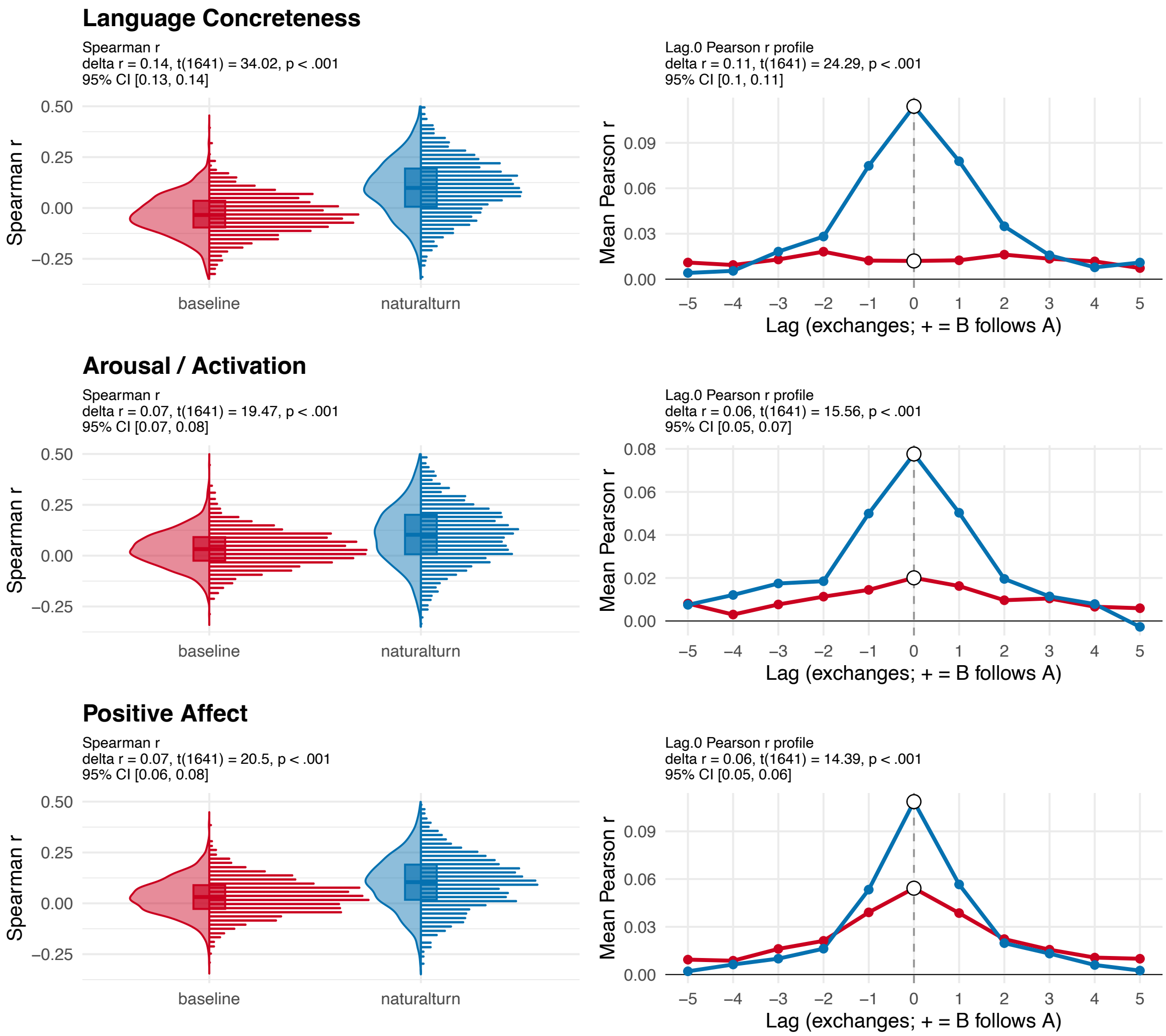

*Note.* Alignment synchrony metrics for three linguistic dimensions (Language Concreteness, Arousal/Activation, Positive Affect). Left panels: Raincloud plots show the distribution of Spearman correlations between speaker time-series for each conversation. Higher values indicate stronger synchrony; speakers tracking each other more closely moment-to-moment. NaturalTurn significantly increases correlations across all dimensions. Right panels: Lagged Pearson correlations (averaged across conversations) reveal the temporal structure of the coupling. Each point shows how strongly speaker B at one exchange correlates with speaker A from earlier or later exchanges. The key finding is at lag 0 (haloed points and dashed lines), NaturalTurn produces sharp, high peaks indicating strong immediate turn-by-turn coupling, while Baseline shows shallow, diffuse patterns spread across multiple turns. Together, these results suggest that naturalistic turn segmentation may capture the fine-grained reciprocity and moment-to-moment coordination of dialogue more faithfully than baseline segmentation.

---------------------------------

## Discussion

NaturalTurn is an algorithm for segmenting transcripts into naturalistic, psychologically meaningful turns. When deployed on a large conversation corpus, the NaturalTurn algorithm produced transcripts with turn durations and intervals that were seemingly more representative of a conversation's actual flow compared to baseline transcripts. NaturalTurn-generated transcripts also provide evidence for a distinct relationship between turn duration and a range of subjective impressions, including people's affect, enjoyment, and sense of shared reality with their conversation partner, while analyses that relied on the baseline transcripts showed none of these results. Finally, NaturalTurn produced segmented transcripts that revealed stronger, more nuanced, and intuitively plausible patterns of alignment, convergence, and synchrony between conversation partners on several dimensions such as word concreteness, arousal/activation, and positive affect. By relating micro-level turn dynamics and macro-level outcomes that result from social interaction—an increasingly common and important type of research in the social and computational sciences—NaturalTurn represents a methodological improvement for researchers seeking to perform various types of conversation analysis at corpus scale, without laborious annotation, feature extraction, or additional machine-learning infrastructure.

### Contributions & Limitations

NaturalTurn's simple heuristic (i.e., the floor stays with the current speaker until their silence exceeds a parameterized maximum) is not without limitations. Inevitably, NaturalTurn will incorrectly label some proportion of legitimate primary turns as secondary speech, as well as incorrectly label some portion of secondary turns as primary turns. A first line of improvement may simply be to tune the critical silence threshold (max_pause), the backchannel cue list, and other parameters. Further advancements could involve refining the heuristics used here or adding new ones to better capture the nuances of conversational flow, including multimodal techniques that can predict turn endings by taking advantage of additional acoustic and visual cues. Finally, in an ideal scenario, a sufficiently large dataset with turns labeled by the participants themselves (to reflect their subjective experience of who actually held the floor at each moment) would pave the way for a trained model to recognize primary turns directly, bypassing the constraints of NaturalTurn's heuristic-based architecture. Given these considerations, NaturalTurn represents a preliminary, and largely pragmatic, solution to a complex challenge, one that will benefit from the collaborative efforts of multidisciplinary research teams in the fields of psychology, communications, linguistics, and machine intelligence.

Parallel speech is a common feature of natural conversation, which NaturalTurn was designed to analyze. However, NaturalTurn's utility diminishes in social exchanges that lack parallel speech, as in highly structured meetings, for example. Essentially, if there is no parallel speech, NaturalTurn has limited effect on a transcript. However, the more parallel the speech, there is in a transcript—listeners' ancillary intrusions into the main conversational flow—the more effective NaturalTurn becomes.

Beyond its core function related to turn segmentation, NaturalTurn's accessory features enhance its functionality and ease of use, such as the ability to flag and/or remove low-confidence tokens and known mistranscriptions. For a comprehensive view of NaturalTurn's capabilities, see its source code and the full set of transcripts for the entire CANDOR corpus: https://github.com/betterup/natural-turn-transcription; https://osf.io/xy35p/. We encourage researchers to optimize such features alongside the basic segmentation functionality.

At present NaturalTurn assigns any brief listener utterance that passes the lexical cue tests to a single *Backchannel* class and labels everything else that intrudes on the floor as an undifferentiated *Secondary Turn*. This coarse taxonomy is a convenience, not a theoretical claim. In fact, the backchannel category itself conflates several functions long distinguished in the literature: minimal continuers that simply license the speaker to proceed ("uh-huh," "mhm"), affiliative assessments that mirror the speaker's stance ("exactly," "wow"), and turn-shaping signals such as news receipts or sequence-closing tokens [11-13]. Likewise, the residual pool of secondary turns surely contains a mix of mis-segmented primary talk, overlooked backchannels, and other listener actions that deserve their own labels. Future work can therefore proceed on two fronts. First, the existing *Backchannel* class can be split by supplementing the cue list with functional sub-lists or by clustering on prosodic and lexical features. Second, the remainder of the *Secondary Turn*s can be audited, both automatically and by hand, to estimate how many events are mis-classified primaries versus novel categories of parallel speech, thereby guiding the introduction of additional, better-defined labels. That iterative refinement will turn NaturalTurn's current catch-all tiers into a richer inventory of listener behaviors and may even surface new phenomena that have been difficult to study at scale.

Regarding generalizability, NaturalTurn is essentially language-agnostic because it requires only word-level timestamps attached to each speaker, outputs that most modern ASR systems provide for many languages. Porting the tool therefore involves tweaking two tunable parameters: backchannel lexicon and floor-release max_pause parameter. The backchannel lexicon filter relies on a short list of listener-feedback tokens (e.g., mhm, yeah). Researchers can replace this list with counterparts in the target language or dialect (e.g., un, ee, hai), leaving the code unchanged. The max_pause threshold that signals a speaker has yielded the floor may need retuning to match typical gap lengths in the new linguistic or interactional context. Both parameters live in plain-text configuration files and can be edited without touching the algorithm.

While a new language makes these adjustments salient, the same fine-tuning is advisable for regional variants of English, different media (phone vs. face-to-face), interaction types (task-oriented vs. affiliative), or other corpora collected in distinct social settings.

NaturalTurn is specifically designed for dyadic conversations and does not currently work for group interactions with three or more speakers. While the archetype of conversation may be between two people, a considerable portion of our lives is spent talking in groups, where everything we have explored (turn-taking, parallel speech, conversational coordination) has the potential to change when more voices, minds, and preferences are combined. Unfortunately, these differences, which one can feel intuitively, have been scarcely examined [82-84]. Group conversations present even more complex challenges for managing parallel speech, as multiple listeners may simultaneously provide feedback, compete for the floor, or engage in side conversations. Yet this complexity also represents an exciting unexplored frontier. We hope some intrepid researchers will take up the challenge of adapting and extending approaches like NaturalTurn to investigate how group conversations should best be segmented into turns, a critical step on the path to understanding the nuanced social dynamics that characterize multi-party conversation.

## Outlook & Future Work

Two fundamental features of turn-taking dynamics are the duration of turns and the intervals between them. Our main emphasis here has been on turn duration because in the decades since the earliest investigations [14], many streams of research have commenced, only to stall, in part because of the lack of automated methods to segment speech into naturalistic turns. Looking forward, there is immense potential to explore a wider array of turn-taking features, such as minimum and maximum turn lengths, how turn durations vary within a conversation, and the synchrony of turns between conversation partners [16-17]. With the advent of new turn models, such as NaturalTurn, we suspect that researchers will be better able to model patterns of speech durations and associate these dynamics with meaningful social outcomes.

In contrast to turn duration, the literature on turn intervals is comparatively well-developed. A central finding is the remarkably brief 100–200 ms mean gap between turns, which places strong constraints on potential underlying cognitive mechanisms (i.e., listeners must be predicting the end of speakers' turns to take over the floor with such precision) [9, 74, 85]. Ostensibly, NaturalTurn improves on baseline transcripts because its mean turn interval is more consistent with what is reported in the literature. That said, it is worth noting that NaturalTurn was not optimized for interval analysis in this study; and indeed, in our testing, the silence threshold we chose significantly impacted interval dynamics, as did the specific speech-to-text service chosen. Thus, while NaturalTurn's potential to large-scale data on turn intervals is an exciting prospect, additional development is required.

When correlating turn dynamics with conversation outcomes, this study focused necessarily on a very limited set of analyses, i.e., the simple correlation between turn duration and people's emotional reactions to their conversation overall. However, given the ability to extract naturalistic turns at scale, research is poised to relate turn-taking to an even more diverse array of conversational outcomes, such as developing relationships [86], establishing norms [87], negotiating conflict [88], impression formation [89], building social connection [19], and regulating one's mental and physical health [90, 91].

The utility of turn models extends far beyond accurately representing the duration and intervals between turns, as they can also aggregate other types of conversational data. As technology increasingly allows for the extraction of facial emotion, measures of semantic distance, vocal prosody, and so forth, these observations need to be organized effectively, often around the unit of a turn. By effectively segmenting turns, NaturalTurn creates psychologically coherent "containers" for all such data, forming the basic conversational unit analyzable for a range of empirical inquiries. Our alignment/synchrony case study illustrates the point: when word concreteness, arousal, and affect were aggregated over NaturalTurn boundaries, the partner-to-partner coupling was stronger, more theoretically plausible, and showed patterns unanticipated in the baseline turn model. Any phenomenon that depends on "who was speaking when?" can benefit in exactly the same way once its signals are folded into NaturalTurn's segmentation structure.

The evolution of turn models could eventually yield an automated standard for organizing psychologically meaningful, naturalistic turns. The proliferation of new models presents many opportunities for collaborative development, but also will require thoughtful management of the research variance introduced by the many different ways to segment transcripts and aggregate data in various domains. These "researcher degrees of freedom" must be managed thoughtfully. Indeed, one of NaturalTurn's advantages is that its flexible, parameterized algorithm allows researchers to encode their working assumptions and report these choices clearly. In this spirit, we envision future directions for turn model development that will not only refine model quality, but also establish clear usage and reporting guidelines, thereby bolstering the reliability and scientific integrity of conversational analysis.

The analysis of turn-by-turn conversation holds significant promise as a basic and applied research enterprise. Despite its importance, researchers lack automated, annotation-free, and scalable methods to segment transcripts into psychological meaningful conversational turns.

NaturalTurn addresses this gap by providing researchers and practitioners with a tool that more accurately reflects the naturalistic structure of dialogue, inviting new investigation into how conversational dynamics are related to a broad range of social outcomes.

## Data, Code, & Algorithm Availability

All data and analysis scripts needed to evaluate the conclusions in this paper are available. The authors are also releasing the full NaturalTurn algorithm publicly for the broader scientific community.

NaturalTurn open-source code is available at:
https://github.com/betterup/natural-turn-transcription

Data, code, and analysis scripts can be found at:
https://osf.io/xy35p/

## Funding & Acknowledgments

The authors wish to thank BetterUp, Inc. for its willingness to share the details of this algorithm with the wider scientific community, Auriane Boudin for her insightful comments on the manuscript, and ELC for the helpful edits.

## Competing Interests Disclosure

At the time this project was conducted, AR was an employee, and GC was a consultant at BetterUp Inc. The authors declare that they have no other competing interests.

## Author Contributions

GC and AR share first authorship.

Conceptualization: GC, AR

Methodology: GC, AR

Investigation: GC, AR

Formal analysis: GC

Visualization: GC

Writing – original draft: GC

Writing – review & editing: GC, AR